\documentclass{article}
\usepackage{arxiv}
\usepackage[utf8]{inputenc} 
\usepackage[T1]{fontenc}    
\usepackage{hyperref}       
\usepackage{url}            
\usepackage{booktabs}       
\usepackage{amsfonts}       
\usepackage{nicefrac}       
\usepackage{microtype}      
\usepackage{lipsum}		

\usepackage{algcompatible} 
\usepackage{graphicx}
\usepackage[round,authoryear]{natbib}
\usepackage{doi}
\usepackage{amsmath}
\usepackage{amsthm}
\usepackage{tcolorbox}
\usepackage{xcolor}
\usepackage{courier} 
\usepackage{amssymb}
\usepackage{listings}
\usepackage{enumitem}
\usepackage[ruled]{algorithm2e}

\SetKwInOut{KwIn}{Input}
\SetKwInOut{KwOut}{Output}

\usepackage{wrapfig}
\usepackage{multirow}

\usepackage{caption}
\definecolor{promptbg}{RGB}{252,252,255}
\definecolor{promptborder}{RGB}{70,130,180}
\definecolor{prompttext}{RGB}{20,20,60}
\definecolor{promptheader}{RGB}{70,130,180}
\definecolor{codebg}{RGB}{250,250,250}
\definecolor{codegray}{RGB}{128,128,128}

\tcbuselibrary{skins,breakable}

\newtcolorbox{promptbox}[1][]{
    colback=promptbg,
    colframe=promptborder,
    boxrule=2pt,
    arc=4pt,
    breakable,
    enhanced,
    before skip=12pt,
    after skip=12pt,
    left=8pt,
    right=8pt,
    top=8pt,
    bottom=8pt,
    fontupper=\fontfamily{pcr}\selectfont\footnotesize,
    shadow={2pt}{-2pt}{0pt}{black!15},
    title={#1},
    coltitle=white,
    colbacktitle=promptborder,
    fonttitle=\bfseries,
    attach boxed title to top left={yshift=-2mm, xshift=4mm},
    boxed title style={
        arc=2pt,
        boxrule=0pt,
    }
}
\lstdefinestyle{promptstyle}{
    backgroundcolor=\color{codebg},
    basicstyle=\ttfamily\small,
    breakatwhitespace=false,
    breaklines=true,
    keepspaces=true,
    numbers=none,
    showspaces=false,
    showstringspaces=false,
    showtabs=false,
    tabsize=2,
    frame=single,
    rulecolor=\color{promptborder}
}

\newtheorem{theorem}{Theorem}
\theoremstyle{plain}
\newtheorem{definition}{Definition}

\title{\textsc{BarrierBench:} Evaluating Large Language Models for Safety Verification in Dynamical Systems}
\author{}

\author{%
\textbf{Ali Taheri}$^{1}$ \quad
\textbf{Alireza Taban}$^{1}$ \quad
\textbf{Sadegh Soudjani}$^{2}$ \quad
\textbf{Ashutosh Trivedi}$^{3}$
}
\date{}

\renewcommand{\shorttitle}{BarrierBench: Evaluating LLMs for Safety Verification in Dynamical Systems}

\hypersetup{
pdftitle={A template for the arxiv style},
pdfsubject={q-bio.NC, q-bio.QM},
pdfauthor={David S.~Hippocampus, Elias D.~Striatum},
pdfkeywords={First keyword, Second keyword, More},
}

\begin{document}
\maketitle
\renewcommand{\thefootnote}{}
\footnotetext{$^1$ Isfahan University of Technology, Iran. (Email:  \{taheri.a, taban.a\}@ec.iut.ac.ir) $^2$ Max Planck Institute for Software Systems, Germany. (Email: sadegh@mpi-sws.org) $^3$ University of Colorado Boulder, USA. (Email: ashutosh.trivedi@colorado.edu)}

\begin{abstract}
Safety verification of dynamical systems via barrier certificates is central to ensuring correctness in autonomous and other safety-critical applications. Synthesizing such certificates requires discovering mathematical functions that characterize inductive state invariants and provably separate safe and unsafe regions. Existing approaches, however, often struggle to scale computationally, depend on carefully designed templates, and rely on exhaustive or incremental searches over function spaces. They also demand substantial manual ingenuity and mathematical sophistication in constructing the search infrastructure, including selecting template families, choosing appropriate solvers, tuning hyperparameters in data-driven methods, and designing effective sampling procedures.

As a result, successful barrier certificate synthesis requires both a deep understanding of dynamical systems and control theory and practical experience with existing synthesis techniques. Much of this expertise has traditionally been transmitted among practitioners through natural language rather than formalized mathematical procedures.

This observation raises a natural question: can the linguistic and analogical reasoning that experts use informally be captured and operationalized by large language models (LLMs)? Motivated by this question, we present an \emph{LLM-agentic framework for barrier certificate synthesis} that uses natural language reasoning to propose, refine, and validate candidate certificates. The framework combines \emph{LLM-driven template discovery} with \emph{SMT-based verification} and supports \emph{barrier-controller co-synthesis} for controlled systems, thereby ensuring mathematical compatibility between safety certificates and feedback control laws.

To evaluate this capability, we introduce \emph{BarrierBench}, a benchmark of 100 dynamical systems spanning linear, nonlinear, discrete-time, and continuous-time settings, including 68 controlled systems that require barrier-controller co-synthesis. Our experiments assess not only the effectiveness of LLM-guided barrier synthesis but also the value of \emph{retrieval-augmented generation (RAG)} and \emph{agentic coordination strategies} in improving reliability and performance. Across these tasks, the framework achieves over 90\% success in generating valid certificates and demonstrates structural diversity, ranging from simple quadratic forms to high-order coupled polynomials. By releasing \textsc{BarrierBench} with the accompanying toolchain, we aim to establish a \emph{community testbed} for advancing the integration of language-based reasoning with formal verification for dynamical systems. The benchmark is publicly available at \url{https://hycodev.com/dataset/barrierbench}.
\end{abstract}


\section{Introduction}
\label{sec:introduction}
This century has witnessed an increasingly sophisticated integration of digital software with physical infrastructure, interconnected through wired and wireless networks. These cyber-physical systems (CPS) operate so seamlessly and intuitively that it is easy to overlook the complexity beneath their surface: their successful operation depends on intricate control logic whose correctness requires a deep understanding of both the underlying physics and the discrete switching logic governing their behavior. As technological practice continues to deliver critical infrastructures such as autonomous vehicles, implantable medical devices, and smart grids, ensuring their safety has become an increasingly demanding challenge for control engineers amid ever-growing system complexity. 
In this paper, we address this challenge by exploring how large language models (LLMs) can assist in the synthesis and verification of safety guarantees for complex CPS.

\smallskip \noindent \textbf{Safety verification of dynamical systems} is a canonical problem in the design of safety-critical CPS. Since safety verification is computationally undecidable even for CPS with simple piecewise-constant dynamics~\citep{asarin1998achilles}, fully automatic procedures for obtaining provable safety guarantees remain out of reach. Consequently, a variety of methods has been developed to construct safety proofs, including symbolic state-space exploration, statistical guarantees, and abstraction-based techniques. Among these, the synthesis of \emph{barrier certificates}~\citep{prajna2004safety,ames2019control,jagtap2020formal} has emerged as a particularly effective and scalable approach. Barrier certificates serve as functional analogs of inductive state invariants, characterizing the separation between safe and unsafe regions of the state space. Originating in Lyapunov’s classical theory of stability~\citep{lyapunov1966stability,khalil2002nonlinear}, they extend the central idea of certifying system behavior from convergence to invariance. Their formulation as the search for a single scalar function satisfying suitable inequality constraints makes them amenable to optimization- and satisfiability-based methods, contributing to their prominence in both theory and practice. We therefore focus on barrier-certificate-based proof discovery as a promising foundation for integrating formal reasoning with language-model-guided synthesis.

\smallskip
\noindent \textbf{A barrier certificate} is a scalar function whose level sets define an inductive invariant separating unsafe states from all trajectories originating from admissible initial conditions. By providing formal guarantees over infinite time horizons, barrier certificates are central to correctness arguments for both linear and nonlinear dynamical systems. Traditional approaches to barrier synthesis are predominantly optimization-based. Sum-of-squares and semidefinite programming methods cast the search as a convex optimization problem over polynomial templates~\citep{prajna2007framework,kordabad2025sum}. These methods provide strong correctness guarantees, but their computational cost grows rapidly with system dimension, and they are typically limited to low-degree polynomial classes. To improve generality, counterexample-guided inductive synthesis (CEGIS) frameworks---see \cite{dawson2023safe} for a comprehensive review---combine such methods with Satisfiability Modulo Theories (SMT) solvers to iteratively refine candidate barriers~\citep{ravanbakhsh2015counter}. Neural and data-driven variants use deep architectures to approximate more expressive certificate functions~\citep{abate2024safe,neustroev2025neural,abate2021fossil}. Although these approaches broaden the space of representable barriers, they also introduce new challenges, including training instability, reliance on finite samples, reduced interpretability, and continued dependence on carefully chosen templates or architectures. As a result, successful synthesis still depends heavily on expert intuition in selecting suitable function classes, solvers, and hyperparameters. These decisions often reflect informal reasoning that remains difficult to encode within existing algorithmic frameworks. This paper investigates whether LLMs can capture and operationalize the linguistic and analogical reasoning that practitioners use when constructing barrier certificates.

\smallskip
\noindent \textbf{We present an LLM-guided synthesis framework} for barrier certificates that leverages natural-language reasoning to propose, refine, and validate candidate certificates. The framework combines \emph{LLM-driven template discovery} with \emph{SMT-based formal verification} and naturally extends to \emph{barrier--controller co-synthesis} for controlled systems. By incorporating verification feedback into subsequent LLM prompts, the framework supports dynamic exploration of barrier templates and coefficients, often uncovering novel mathematical forms that fixed-template approaches~\citep{prajna2007framework,kong2013exponential,ames2019control,abate2021fossil,kordabad2025sum} may overlook. LLMs guide the synthesis process by reasoning about system properties, safety requirements, and mathematical structure, first analyzing local safety conditions and then integrating them into global certificates. This integration provides the expressiveness and adaptability needed for complex systems while preserving the mathematical rigor required in safety-critical applications.

\begin{figure}[t!]
    \centering
    \includegraphics[width=0.8\textwidth]{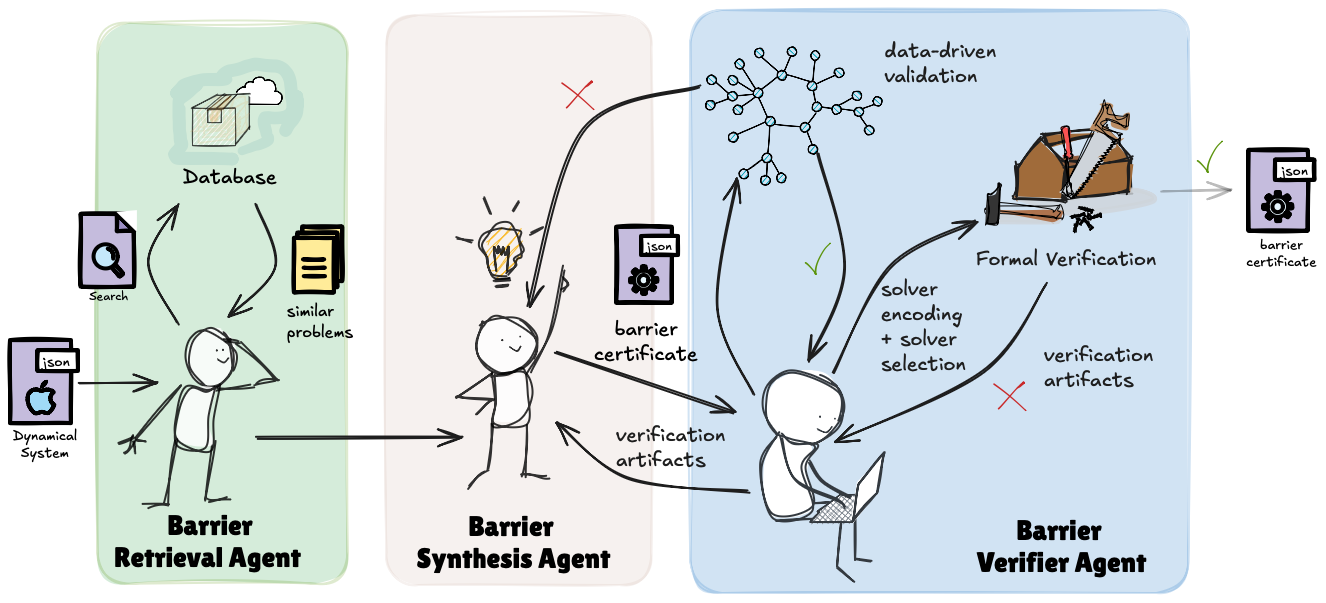}
    \caption{Agentic framework for LLM-guided barrier certificate synthesis.}
    \vspace{-1em}
    \label{fig:framework-architecture}
\end{figure}

To operationalize this idea, we design an \emph{agentic architecture} that emulates the workflow of expert control designers. Human designers typically draw on prior experience with similar dynamical systems to select barrier templates, solver strategies, and proof heuristics. Analogously, our framework instantiates three collaborating agents: a \emph{Barrier Retrieval Agent}, which searches a database of previously solved systems to identify analogous examples; a \emph{Barrier Synthesis Agent}, which proposes candidate barrier certificates inspired by retrieved analogs; and a \emph{Barrier Verifier Agent}, which evaluates these candidates in two stages, first through a lightweight data-driven check and then through formal SMT-based verification. Feedback from the Verifier Agent is returned to the Synthesis Agent, forming a closed-loop interaction bounded by a global timeout. This design enables iterative refinement of candidate certificates while making effective use of external formal tools. Our study investigates the following research questions:
\begin{enumerate}[itemsep=0pt, topsep=0pt, parsep=0pt, partopsep=0pt]
\item[\textbf{RQ1:}] Are state-of-the-art large language models already capable of independently designing valid barrier certificates for safety verification tasks?
\item[\textbf{RQ2:}] Does knowledge of barrier certificates for similar systems improve synthesis performance, and can retrieval-augmented generation (RAG) facilitate this transfer?
\item[\textbf{RQ3:}] Can integrating specialized agents---retrieval, synthesis, and verification---into an agentic framework that interfaces with external solvers improve the efficiency and reliability of LLM-guided barrier synthesis?
\end{enumerate}

\smallskip
\noindent \textbf{To evaluate these questions, we introduce \emph{BarrierBench},} a benchmark of 100 dynamical systems spanning linear, nonlinear, discrete-time, and continuous-time settings, including 68 controlled systems that require barrier--controller co-synthesis. Across these tasks, the LLM framework achieves over 90\% success in generating valid certificates and exhibits structural diversity ranging from simple quadratics to high-order coupled polynomials. By releasing \textsc{BarrierBench} and the accompanying toolchain, we aim to establish a community testbed for advancing the integration of language-based reasoning with formal verification for dynamical systems.

\section{Barrier Certificates}
\label{sec:preliminary}
A dynamical system is defined as a tuple $\mathcal{S} = (X, f, X_I, X_U)$, where $X \subseteq \mathbb{R}^n$ is the state space, 
$f: X \rightarrow \mathbb{R}^n$ is a vector field, $X_I \subseteq X$ is the set of initial states, 
and $X_U \subseteq X$ is the set of unsafe states with $X_I \cap X_U = \emptyset$.
For continuous-time systems, the evolution is governed by
\begin{equation}
\dot{x}(t) = f(x(t)), \quad x(0) \in X_I,
\label{eq:dynamical_system}
\end{equation}
where $x(t) \in X$ denotes the state at time $t\ge 0$.
For discrete-time systems, the evolution is given by
\begin{equation}
x[k+1] = f(x[k]), \quad x[0] \in X_I,
\label{eq:discrete_dynamical_system}
\end{equation}
where $k \in \mathbb{N}$ is the discrete time step.

\begin{definition}[Safety Specification]
Given a dynamical system $\mathcal{S} = (X, f, X_I, X_U)$,
$\mathcal{S}$ is \emph{safe} if trajectories starting in $X_I$ remain in $X \setminus X_U$ for all time.
\end{definition}

\begin{definition}[Barrier Certificate]
A \emph{barrier certificate} is a scalar function $B: X \rightarrow \mathbb{R}$ satisfying:
\begin{align}
B(x) &\leq 0, & \forall x \in X_I, \label{eq:initial_condition} \\
B(x) &> 0,  & \forall x \in X_U, \label{eq:unsafe_condition} \\
\nabla B(x) \cdot f(x) &< 0, & \forall x \in X \text{ s.t. } B(x) = 0 \text{ (continuous-time)}, \nonumber\\
\text{ or } B(f(x)) - B(x) &\leq 0, & \forall x \in X \text{ (discrete-time)}. \label{eq:invariance_condition_discrete}
\end{align}
Here, $\nabla B(x) \cdot f(x)$ denotes the Lie derivative $\mathcal{L}_f B(x)$ of a continuously differentiable function $B(x)$ 
along the vector field $f(x)$, defined as 
$\mathcal{L}_f B(x) = \sum_{i=1}^n \frac{\partial B}{\partial x_i} f_i(x)$.
\end{definition}

\begin{theorem}[Safety via Barrier Certificates~\citep{prajna2004safety}] 
\label{thm:barrier_safety} 
If there exists a barrier certificate $B(x)$ for $\mathcal{S}$, then the system $\mathcal{S}$ is safe.
\end{theorem}

\noindent 
The above definitions and theorem are stated for systems without control inputs. If the system has control input $u\in U$, the objective is to design a state feedback law that gives $u$ as a function of $x$ and makes the system safe. For this, the barrier conditions are quantified existentially over the input.



\begin{algorithm}[t]
\caption{Agentic Barrier Certificate Synthesis}
\label{alg:main_framework}
\KwIn{System $(X,f, X_I, X_U)$; max iterations $K$; refinements $R$; input set $U$ for controlled systems}
\KwOut{Barrier $B^*$, controller $C^*$ for controlled systems, or failure}

Initialize: best\_score $\gets 0$; $B^*, C^* \gets \text{null}$

\For{$k \gets 1$ \KwTo $K$}{
  {$\text{e} \gets \textsc{RetrievalAgent}(f, X_I, X_U, \mathcal{D})$;
$(B_k, C_k) \gets \textsc{SynthesisAgent}(\text{e},\textsf{feedback}_k)$}

  \If{\textsc{VerifierAgent.SampleCheck}$(B_k, C_k)$ fails}{
    \textbf{continue}
  }

  $f' \gets f$ with $C_k$;
  $(\textsf{valid}_k, \textsf{feedback}_k) \gets \textsc{VerifierAgent.FormalCheck}(B_k, f')$

  \If{\emph{$\textsf{valid}_k$}}{
    \textsc{RetrievalAgent.store}$(B_k, C_k)$
    \Return{$(B_k, C_k)$}
  }

  \For{$r \gets 1$ \KwTo $R$}{
    $(B_k^r, C_k^r) \gets \textsc{SynthesisAgent.refine}(\textsf{feedback}_k, r)$

    \If{\textsc{VerifierAgent.SampleCheck}$(B_k^r, C_k^r)$ fails}{
      \textbf{continue}
    }

    $f' \gets f$ with $C_k^r$;
    $(\textsf{valid}_k^r, \textsf{feedback}_k^r) \gets \textsc{VerifierAgent.FormalCheck}(B_k^r, f')$

    \If{\emph{$\textsf{valid}_k^r$}}{
      \textsc{RetrievalAgent.store}$(B_k^r, C_k^r)$
      \Return{$(B_k^r, C_k^r)$}
    }
  }
}
\Return{Best partial solution}
\end{algorithm}

\section{Agentic Framework for Barrier Synthesis}
\label{sec:methodology}

Our synthesis framework employs a multi-agent architecture in which specialized LLM-powered agents collaborate to discover barrier certificates. The framework comprises three core agents operating in an iterative pipeline:  
(1) a \emph{Barrier Retrieval Agent} that identifies relevant prior solutions;  
(2) a \emph{Barrier Synthesis Agent} that generates candidate barrier certificates; and  
(3) a \emph{Barrier Verifier Agent} that checks candidates and provides feedback for refinement.  
This iterative process addresses limitations of conventional approaches by enabling informed template selection and adaptive exploration of diverse mathematical structures through verification-guided synthesis.

\paragraph{Framework Pipeline.}
The framework $\mathcal{F}$ takes as input a dynamical system  $(X, f, X_I, X_U)$, with optional control input $U$ for controlled systems. A persistent database stores previously verified barrier certificates, which the Retrieval Agent accesses to accelerate synthesis on similar problems. The pipeline consists of four stages: (1) historical retrieval, (2) LLM-based synthesis, (3) two-stage verification, and (4) iterative refinement guided by feedback.  
The system performs up to $K$ main iterations, each exploring new template structures, and up to $R$ refinements per iteration to adjust coefficients or terms. The process terminates upon discovery of a valid barrier certificate or exhaustion of all attempts. If no valid certificate is found, the framework returns the candidate maximizing the proportion of satisfied conditions. Figure~\ref{fig:framework-architecture} illustrates the overall architecture of $\mathcal{F}$.

\paragraph{Barrier Retrieval Agent.}
This agent analyzes the input system and retrieves relevant examples from the database through a two-stage process.
First, it performs feature-based retrieval using automatically extracted system attributes such as dimension, time domain, linearity, and topology. Candidate problems matching all features are shortlisted.
Second, an LLM-based reasoning stage ranks these candidates for structural similarity, providing the most compatible example to the Synthesis Agent while noting that adaptation may be required.
When no match is found, the agent instructs the Synthesis Agent to proceed independently.
As shown in Section~\ref{subsec:ablation_study}, this retrieval step significantly accelerates convergence by enabling frequent first-try success.

\paragraph{Barrier Synthesis Agent.}
This agent leverages the reasoning abilities of LLMs to analyze system dynamics and propose barrier certificates that satisfy the safety conditions~\eqref{eq:initial_condition}--\eqref{eq:invariance_condition_discrete}.  
During the first main iteration ($k = 1$), it incorporates relevant historical examples from the Retrieval Agent. For later iterations and refinements, it integrates feedback from the Verifier Agent to modify structure and coefficients.  
For systems requiring control synthesis, the agent jointly produces both the barrier certificate and controller expressions to ensure compatibility between safety and control objectives.

\paragraph{Barrier Verifier Agent.}
The Barrier Verifier Agent conducts a two-stage evaluation of candidate barriers.  
In the first stage, it performs a sample-based validation across $N$ points drawn from $X_I$, $X_U$, and $X$, where $N$ is selected to balance computational efficiency with adequate state space coverage; we use $N=5000$ in our experiments to achieve this trade-off, discarding invalid barriers that violate any of the three conditions on at least one of sampled points, to avoid unnecessary SMT solver calls.  
In the second stage, it performs formal verification using SMT solvers such as Z3~\citep{demoura2008z3}, Yices~\citep{dutertre2006yices}, and cvc5~\citep{barbosa2022cvc5}, checking the unsatisfiability of negated barrier conditions.  
The agent selects solvers adaptively based on problem structure (e.g., polynomial vs. transcendental terms), expression complexity, and implicit knowledge of solver capabilities (e.g., whether transcendental functions are supported), and then applies symbolic conversions or Taylor approximations when necessary, and handles verification failures through solver switching or adaptive timeout adjustments.

\paragraph{Iterative Refinement Mechanism.}
When verification fails, the Verifier Agent provides the violated conditions and counterexamples as feedback to the Synthesis Agent, which uses them to refine the barrier.  
In early refinements ($r \leq 2$), the agent adjusts coefficients while preserving structure; in later refinements ($r > 2$), it performs structural modifications such as adding or removing coupling terms while maintaining polynomial degree.  
This strategy balances exploitation (fine-tuning promising templates) and exploration (structural variation across iterations).  
The Verifier Agent logs all intermediate results, enabling cumulative feedback that helps the Synthesis Agent learn from previous attempts.  
As shown in Section~\ref{subsec:ablation_study}, this mechanism substantially increases success rates, with a large number of cases solved within one or two main iterations.

\section{Barrierbench: Experiments and Case Studies}
\label{sec:case_studies}

The complete set of prompts used in our framework are included in the Appendix~\ref{sec:appendix_prompts}.

\subsection{Experimental Setup}
The framework operates with carefully chosen hyperparameters validated across various systems. We set maximum main iterations $K=5$ and maximum refinements per iteration $R=4$. The LLM utilizes Claude Sonnet 4 and ChatGPT-4o with engineered prompts designed to synthesize mathematical expressions while maintaining formal correctness. All experiments use Z3 version 4.15.1, cvc5 version 1.3.1, and Yices version 2.6.4 as the SMT solvers for formal verification. For the verifier agent, we use in total $N=5000$ samples with 30\% from the initial set, 30\% from the unsafe set, and 40\% from the state space.

\subsection{Case Studies}


\begin{table}[t]
\vspace{-10pt}
\centering
\caption{Dataset characteristics of \textsc{BarrierBench}.}
\label{tab:benchmark_statistics}
\begin{tabular}{ll}
\toprule
\textbf{Characteristic} & \textbf{Description} \\
\midrule
Dimensions & 1D–8D \\
Initial set topology & Ball, rectangle \\
Unsafe set topology & Ball, rectangle, union of rectangles \\
System types & Controlled, autonomous \\
Time domain & Continuous-time, discrete-time \\
Dynamics types & Linear, nonlinear \\
\bottomrule
\end{tabular}
\vspace{-10pt}
\end{table}

We evaluate our agentic barrier synthesis framework on \textsc{BarrierBench}, a benchmark comprising 100 dynamical systems. Our evaluation case studies encompasses continuous-time systems, discrete-time systems, and controlled systems requiring simultaneous barrier-controller co-synthesis, with a variety of complex nonlinear dynamics across dimensions from 1D to 8D. \textsc{BarrierBench} includes 68 controlled systems for which our framework demonstrates the simultaneous synthesis of barrier certificates and controllers parameters, which together establish safety guarantees for closed-loop dynamics. The characteristics of our benchmark are shown in Table~\ref{tab:benchmark_statistics}.

\subsection{Qualitative Results}
In this section, we qualitatively review the results of some of the benchmarks synthesized by our framework, for more case studies you can check \url{https://hycodev.com/dataset/barrierbench}.
\begin{itemize}
    \item \textbf{Continuous-Time Systems.}
The 4D system $\dot{x}_1 = -x_1 + 4.5, \dot{x}_2 = x_1 - x_2 - 3, \dot{x}_3 = -0.5x_3, \dot{x}_4 = -0.3x_4$ with state space $\mathbb{R}^4$ and polynomial degree of 1 has rectangular initial set $[1.75,2.25] \times [1.75,2.25] \times [-0.1,0.1] \times [-0.1,0.1]$ and rectangular unsafe region $[9,10] \times [9,10] \times [-5,5] \times [-5,5]$. The framework generates the quadratic barrier $B(x) = (x_1-4.5)^2 + (x_2-1.5)^2 + x_3^2 + x_4^2 - 25$ of degree 2.

\item 
\textbf{Discrete-Time Systems.}
The 2D system $x_1[k+1] = x_1[k] + 0.01(-100x_1[k] - x_2[k]), x_2[k+1] = x_2[k] + 0.01(x_1[k] - 100x_2[k])$ with state space $\mathbb{R}^2$ and polynomial degree $d_f = 1$ has rectangular initial set $[0.1,0.4] \times [0.1,0.55]$ and rectangular unsafe region $[0.45,0.5] \times [0.6,1.0]$. The framework generates the quadratic barrier $B(x) = x_1^2 + x_2^2 - 0.5$ of degree 2.

\item 
\textbf{Controlled Systems with Barrier-Controller Co-Synthesis.}
The 4D system $\dot{x}_1 = x_2 + u_0, \dot{x}_2 = -0.95\sin(x_1) - 0.02x_2 + 0.01x_3 + u_1, \dot{x}_3 = -0.1x_3 + 0.02x_1 + u_2, \dot{x}_4 = -0.03x_4 + 0.01x_2 + u_3$ with state space $\mathbb{R}^4$, control space $\mathbb{R}^4$, and non-polynomial dynamics has initial ball with center $(0,0,0,0)$ and radius 0.3, and unsafe set as the complement of the ball with center $(0,0,0,0)$ and radius 2.5. The framework synthesizes controllers $u_0 = -3x_1 - 1.5x_2, u_1 = -3x_2 + 0.8\sin(x_1), u_2 = -3x_3 - 0.05x_1, u_3 = -3x_4 - 0.02x_2$ with quadratic barrier $B(x) = x_1^2 + x_2^2 + x_3^2 + x_4^2 - 4.0$ of degree 2.
\end{itemize}



\subsection{Ablation Study}
\label{subsec:ablation_study}
In this section, we compare our full pipeline against a baseline that uses only a single LLM prompt without any framework support to evaluate our framework. Moreover, to evaluate the individual contributions of our framework components, we conduct ablation experiments. Table~\ref{tab:ablation} presents the comparison between our framework and a single LLM prompt based on the number of solved problems in \textsc{BarrierBench}.

\begin{table}[t]
\centering
\caption{Comparison of baseline (single-prompt) and agentic framework on \textsc{BarrierBench}.}
\label{tab:ablation}
\begin{tabular}{lcccc}
\toprule
\multirow{2}{*}{\textbf{Approach}} & 
\multicolumn{2}{c}{\textbf{Claude Sonnet 4}} & 
\multicolumn{2}{c}{\textbf{ChatGPT-4o}} \\
\cmidrule(lr){2-3} \cmidrule(lr){4-5}
 & \textbf{Success Rate} & \textbf{Solved} & \textbf{Success Rate} & \textbf{Solved} \\
\midrule
Baseline (Single Prompt) & 41\% & 41/100 & 17\% & 17/100 \\
Full Framework & \textbf{90\%} & \textbf{90/100} & \textbf{46\%} & \textbf{46/100} \\
\textit{Improvement} & \textit{+49\%} & \textit{+49} & \textit{+29\%} & \textit{+29} \\
\bottomrule
\end{tabular}
\end{table}




The baseline approach, which relies on a single prompt LLM generation without any barrier retrieval from the historical database, iterative refinement, or failure-guided refinement, successfully solves only 41 problems (41\%) with Claude Sonnet 4 and only 17 problems (17\%) with ChatGPT-4o. In contrast, our full framework achieves substantial improvements on both models: 90\% success rate with Claude Sonnet 4 and 46\% success rate with ChatGPT-4o, representing improvements of +49 and +29 percentage points, respectively. 
This performance gap validates the significant role of our framework components and provides a negative answer to \textbf{RQ1}. The baseline's limited success rates demonstrate that state-of-the-art LLMs cannot independently design valid barrier certificates without structured framework support. To further analyze the distribution of synthesized valid barriers across main iterations and refinements, Table~\ref{tab:heatmap} presents a detailed breakdown showing when and how problems are solved within our framework.

\begin{table}[t]
\centering
\caption{Number of problems solved at each iteration–refinement combination.
The index of $I$ shows the number of iterations and the index of $R$ shows the number of refinements. For example, the number under
$I_1$–$R_0$ denotes the number of cases solved in first-try without refinement, leveraging prior database search for similar problems.}

\label{tab:heatmap}
\begin{tabular}{lcccc}
\toprule
& $\boldsymbol{R_0}$ & $\boldsymbol{R_{1-2}}$ & $\boldsymbol{R_{3-4}}$ & \textbf{Total} \\
\midrule
$\boldsymbol{I_1}$ & 49 & 9 & 5 & 63 (70.0\%) \\
$\boldsymbol{I_2}$ & 10 & 3 & 2 & 15 (16.7\%) \\
$\boldsymbol{I_{3+}}$ & 4 & 2 & 6 & 12 (13.3\%) \\
\midrule
\textbf{Total} & 63 (70.0\%) & 14 (15.6\%) & 13 (14.4\%) & 90 (100\%) \\
\bottomrule
\end{tabular}
\end{table}

\paragraph{Barrier Retrieval Impact.} 
As shown in Table~\ref{tab:heatmap}, our barrier retrieval mechanism, which combines feature-based filtering and LLM selection accelerates convergence. Among the 90 successfully solved problems, 49 (54.4\%) are solved in the first attempt without any refinements ($I_{1}-R_{0}$), compared to 41 total successes from the baseline. This indicates that our intelligent barrier retrieval not only increases the overall success rate but also accelerates solution discovery. Problems that require multiple iterations or fail entirely under the baseline approach are now solved by leveraging patterns from problems with matching properties, which provides a positive answer to \textbf{RQ2}. 

Our retrieval does not transfer certificates directly. It provides structural templates as few-shot examples for the LLM. Similar to few-shot learning~\citep{NEURIPS2020_1457c0d6}, the synthesis agent uses architectural patterns from analogous problems, then generates entirely new coefficients and controllers tailored to the current problem's specific constraints while explicitly instructed not to copy solutions (see Appendix~\ref{sec:appendix_prompts}). For out-of-distribution scenarios where retrieval fails, the framework gracefully degrades to independent synthesis (baseline 41\% capability) with iterative structural exploration. Importantly, as discussed in Section~\ref{sec:methodology}, failures from initial attempts accumulate as feedback, constrain the exploration space by showing what does not work, enabling the synthesis agent to propose progressively better structures in subsequent iterations.

\paragraph{Refinement Impact.}
The refinement mechanism contributes substantially to the framework performance. As shown in Table~\ref{tab:heatmap}, while 63 problems (70\%) require no refinement ($R_0$ column), an additional 14 problems (15.6\%) benefit from non-structural refinement ($R_{1-2}$), and 13 problems (14.4\%) require structural adjustments ($R_{3-4}$). Within the first iteration alone, the refinement mechanism successfully synthesizes certificates for an additional 14 problems beyond the initial 49, bringing {$I_1$} success to 63 solved problems. This demonstrates that even when the initial LLM-generated template requires adjustment, the refinement mechanism can improve coefficients and, when necessary, modify mathematical structures to satisfy all safety conditions.

\paragraph{Multi-Iteration Synthesis Impact.}
As demonstrated in Table~\ref{tab:heatmap}, for problems not solved in the first iteration, the framework's ability to generate structurally different templates in later iterations proves essential, bringing additional 27 successful barrier synthesis. The second iteration ($I_2$) successfully solves 15 additional problems (16.7\%), bringing the cumulative success rate to 86.7\%. The remaining 12 problems (13.3\%) require three or more iterations ($I_{3+}$), with half of these requiring structural refinement ($R_{3-4}$), indicating that more challenging problems often require exploration of alternative mathematical forms, as the proportion of problems requiring structural refinement increases in later iterations (from 7.9\% in $I_1$ to 50\% in $I_{3+}$). 
The results of the ablation study provide a positive answer to \textbf{RQ3}, demonstrating the improvement in the efficiency and reliability of LLM-guided barrier synthesis through the agents interaction.




\paragraph{Limitations.}

Although our synthesis framework outperforms the baselines by solving 90\% of the \textsc{BarrierBench} problems, the remaining 10 unsolved problems reveal some synthesis challenges. The unsolved cases are highly complex nonlinear dynamics, including systems with complex combination of trigonometric and exponential terms, across different topological sets. Their complexity challenges both the synthesis framework and the verification solvers. In certain cases, the Synthesis Agent fails to generate barrier certificates satisfying all safety conditions due to the complexity of the underlying dynamics.
Moreover, for the synthesized barriers, SMT solvers face computational difficulties during verifying conditions as system dimension and nonlinear complexity increase, frequently timing out or failing to determine satisfiability.

\section{Related Work}
\label{sec:related_work}

\smallskip
\noindent \textbf{Barrier certificate generation.}
Classical approaches to barrier certificate synthesis are predominantly optimization-based, especially sum-of-squares (SOS) programming~\citep{prajna2007framework,ames2019control,wooding2025protect,kordabad2025sum}, but their computational cost grows quickly with system dimension and complexity. To improve scalability, counterexample-guided inductive synthesis (CEGIS) frameworks such as FOSSIL~\citep{abate2021fossil} integrate neural networks with SMT-based verification, enabling iterative refinement of nonlinear barrier certificates beyond fixed polynomial templates. Barrier certificates have also been extended beyond safety to richer specifications, including signal temporal logic~\citep{lindemann2018control}, linear temporal logic over finite traces~\citep{jagtap2020formal}, and reachability specifications~\citep{majumdar2024necessary}. 

When system dynamics are difficult to model exactly, data-driven approaches aim to infer safe behavior from sampled trajectories~\citep{nejati2023formal}. Much of this literature focuses on linear or control-affine dynamics~\citep{Jagtap2020CBCGP,Cohen2022,Lopez2022uCBF}, often under partial knowledge or structural assumptions needed for formal guarantees. For example, \citep{Wang2018CBF} learns unknown dynamics with Gaussian processes while assuming known control-affine structure, and \citep{Jagtap2020CBCGP} assumes known control input while learning the unknown drift. Other works impose assumptions on stochastic components, for example by assuming unknown noise distributions over known deterministic dynamics~\citep{mathiesen2024data} or known noise distributions~\citep{mazouz2024data}. More recent work considers systems with fully unknown dynamics~\citep{Salamati2021DDCBC,wang2023stochastic,schon2024DRObarrier}, supported by tools such as LUCID~\citep{lucid}. Despite these advances, most existing methods still rely on manually chosen templates, motivating more flexible and automated synthesis approaches.

\smallskip
\noindent \textbf{LLMs for complex reasoning.}
Recent LLMs exhibit strong reasoning, planning, and problem-solving abilities~\citep{achiam2023gpt,xu2024llm,zhao2023survey}. Models such as GPT-3~\citep{NEURIPS2020_1457c0d6,dale2021gpt3} and GPT-4~\citep{sanderson2023gpt4} demonstrate broad generalization grounded in large-scale pretraining~\citep{xu2024llm}. They can generate multi-step plans from high-level instructions~\citep{jansen2020visually}, decompose complex problems into subproblems, and reason across multiple levels of abstraction~\citep{lin2023text2motion,rana2023sayplan}. These capabilities have been applied in domains ranging from robotic task planning~\citep{rana2023sayplan} and autonomous driving~\citep{cui2023large} to mathematical reasoning involving symbolic manipulation, quantitative analysis, and formal logic~\citep{lewkowycz2022solving}. This progress makes LLMs promising tools for safety-critical settings that require structured reasoning about system constraints and safe behavior.

\smallskip
\noindent \textbf{LLMs for CPS.}
LLMs are increasingly being explored in CPS as engines for high-level reasoning, planning, and natural-language interaction~\citep{xu2024llm}. Their broad pretraining across language, code, mathematics, and science gives them a wide base of transferable knowledge. In robotics, SayCan~\citep{ahn2022saycan} combines LLM reasoning with reinforcement learning by using affordance-based value functions to ground language commands in executable actions. In autonomous driving, DriveGPT4~\citep{yang2023drivegpt4} uses LLMs for interpretable decision-making, generating natural-language explanations together with low-level control predictions. In formal controller synthesis, \citep{bayat2025llm} propose a multi-agent framework that translates natural-language specifications into verified symbolic control code by coupling reasoning with abstraction-based verification. At the same time, LLMs remain prone to hallucination~\citep{xu2024llm}, lack intrinsic grounding in physical constraints, and are difficult to verify with traditional neural-network verification tools such as Reluplex. These limitations motivate hybrid frameworks that combine LLM-based reasoning with formal verification to ensure correctness and safety in CPS.

\section{Conclusion}
\label{sec:conclusion}

We presented an LLM-agentic framework for barrier certificate synthesis that combines natural-language reasoning, retrieval of relevant prior barriers, and verification-guided refinement to address key limitations of existing methods. The framework automates much of the manual process of template selection, discovers diverse barrier structures beyond fixed-template approaches, and extends to barrier--controller co-synthesis for controlled systems. We also introduced \textsc{BarrierBench}, a benchmark of 100 dynamical systems spanning continuous-time, discrete-time, linear, nonlinear, and controlled settings, on which the framework achieves a 90\% success rate.

This work takes an initial step toward integrating language-based reasoning with formal safety verification and control synthesis. Rather than claiming exhaustive coverage, we aim to provide a concrete and extensible starting point for future work. We hope that releasing both the framework and \textsc{BarrierBench} will support shared experimentation and accelerate progress toward more general, interpretable, and automated methods for safety assurance in dynamical systems.

\section*{Acknowledgements}
This research was supported in part by the National Science Foundation under CAREER Award CCF-2146563. Ashutosh Trivedi is a Royal Society Wolfson Visiting Fellow and gratefully acknowledges support from the Wolfson Foundation and the Royal Society. 
The research of Sadegh Soudjani is supported by ERC 101089047 and by EIC 101070802.
We also acknowledge the use of the Excalidraw tool (\url{https://excalidraw.com}) in creating the illustration in Figure~\ref{fig:framework-architecture}.

\bibliographystyle{plainnat}
\bibliography{references}

\newpage
\appendix

\section{LLM Prompts}
\label{sec:appendix_prompts}

This appendix presents the complete set of prompts used in our framework.

\subsection{Prompt 1: Dataset Similarity Selection}

This prompt selects the most similar problem from filtered candidates using LLM-based similarity assessment.

\begin{promptbox}[Dataset Similarity Selection]
\textcolor{promptheader}{\textbf{TARGET PROBLEM:}}\\
Dynamics: \{SYSTEM\_DYNAMICS\}\\
Initial set: \{INITIAL\_SET\}\\
Unsafe set: \{UNSAFE\_SET\}

\vspace{1.5ex}

\textcolor{promptheader}{\textbf{COMPATIBLE CANDIDATES (all are fundamentally similar):}}\\
\{CANDIDATES\_TEXT\}

\vspace{1.5ex}

\textit{Which candidate has the most similar problem type and structure to the target problem?}

\vspace{1.5ex}

Focus on: system structure, problem type, and mathematical pattern similarity.

\vspace{1.5ex}

Answer with only the candidate number (1, 2, 3, etc.):
\end{promptbox}

\subsection{Prompt 2: Barrier Generation in the First Iteration}

This prompt generates the barrier certificate for systems incorporating the retrieved context if available.

\subsubsection{Systems Without Controller}
\begin{promptbox}[Barrier Synthesis in the First Iteration - No Controller]
\{CONTEXT\}

\vspace{1.5ex}

\textcolor{promptheader}{\textbf{Main Problem:}}\\
- Dynamics: \{SYSTEM\_DYNAMICS\}\\
- Initial set: \{INITIAL\_SET\}\\
- Unsafe set: \{UNSAFE\_SET\}

\vspace{1.5ex}

\textcolor{promptheader}{\textbf{Design a barrier certificate B(x) that satisfies:}}\\
1. B(x) $\leq$ 0 in initial set\\
2. B(x) > 0 in unsafe set\\
3. \{CONDITION\_3\}

\vspace{1.5ex}

Be very careful - don't make it more complex than needed.

\vspace{1.5ex}

\textcolor{red}{\textbf{CRITICAL:}} Use ONLY real numbers in the barrier expression. No variables like
'c' or '$\varepsilon$'.
\vspace{1.5ex}

Solve specifically for THIS problem with appropriate coefficients.

\vspace{1.5ex}

Analyze carefully but be concise. Give precise answer without long explanations.

\vspace{1.5ex}

\textcolor{promptheader}{\textbf{Format your response as (don't make it bold):}}

\vspace{1.5ex}

BARRIER: [expression with numbers only]
\end{promptbox}

The variable \texttt{CONDITION\_3} refers to
\begin{itemize}
    \item Discrete-time: $B(f(x)) - B(x) \leq 0$ for all $x$ in the state space;
    \item Continuous-time: $\nabla B(x) \cdot f(x) < 0$ on the boundary.
\end{itemize}
 
The variable \texttt{CONTEXT} refers to the following two cases:

(a) when a similar problem exists in the dataset:
\begin{lstlisting}[style=promptstyle, basicstyle=\ttfamily\scriptsize]
Related problem found:
EXAMPLE: {SIMILAR_PROBLEM}
B(x): {SIMILAR_BARRIER}

WARNING: This is just an example - your solution may be 
completely different NOT ONLY in terms of coefficients, 
BUT ALSO in format and structure. You may make mistakes, 
so do not consider this as a solution. Please don't copy it.
\end{lstlisting}

(b) when no similar problem exists:
\begin{lstlisting}[style=promptstyle, basicstyle=\ttfamily\scriptsize]
No similar problems found. Analyze this problem fresh.
\end{lstlisting}

\subsubsection{Systems With Control Inputs}

\begin{promptbox}[Barrier and Controller Synthesis in the First Iteration]
\{CONTEXT\}

\vspace{1.5ex}

\textcolor{promptheader}{\textbf{Main Problem:}}\\
- Dynamics: \{SYSTEM\_DYNAMICS\}\\
- Initial set: \{INITIAL\_SET\}\\
- Unsafe set: \{UNSAFE\_SET\}

\vspace{1.5ex}

\textcolor{promptheader}{\textbf{CONTROLLER SYNTHESIS:}} This system has control inputs
\{CONTROLLER\_PARAMETERS\}. 

You need to design BOTH:\\
1. Barrier certificate B(x)\\
2. Controller expressions for \{CONTROLLER\_PARAMETERS\}

\vspace{1.5ex}

The controller u(x) will be substituted into dynamics to create closed-loop system.

\vspace{1.5ex}

\textcolor{promptheader}{\textbf{Design a barrier certificate B(x) that satisfies:}}\\
1. B(x) $\leq$ 0 in initial set\\
2. B(x) > 0 in unsafe set\\
3. \{CONDITION\_3\}

\vspace{1.5ex}

Be very careful - don't make it more complex than needed.

\vspace{1.5ex}

Design both barrier certificate B(x) AND controller expressions that work\\
together to satisfy all conditions.

\vspace{1.5ex}

\textcolor{red}{\textbf{CRITICAL:}}\\
- Use ONLY real numbers in both barrier and controller expressions.\\
  No variables like 'c' or '$\varepsilon$'.\\
- Solve specifically for THIS problem with appropriate coefficients.\\
- Controller must be implementable with realistic actuators\\
- Ensure controller bounds are reasonable (avoid extremely large values)

\vspace{1.5ex}

Analyze carefully but be concise. Give precise answer without long explanations.

\vspace{1.5ex}

\textcolor{promptheader}{\textbf{Format your response as (don't make it bold):}}

\vspace{1.5ex}

BARRIER: [barrier expression with numbers only]\\
CONTROLLER: [controller expressions for each parameter, comma-separated]
\end{promptbox}

\subsection{Prompt 4: Barrier Generation in Subsequent Iterations}

For subsequent iterations, the prompt leverages failure history to improve structure.

\subsubsection{Systems Without Controller} 
\begin{promptbox}[Barrier Synthesis in Subsequent Iterations - No Controller]
\textcolor{promptheader}{\textbf{Previous attempts failed:}}\\
- Tried: \{BARRIER\} \{FAILED\_INFO\}\\
- Tried: \{BARRIER\} \{FAILED\_INFO\}\\
....

\vspace{1.5ex}

\textit{Improve the barrier structure to satisfy all conditions.}

\vspace{1.5ex}

\textcolor{promptheader}{\textbf{Main Problem:}}\\
- Dynamics: \{SYSTEM\_DYNAMICS\}\\
- Initial set: \{INITIAL\_SET\}\\
- Unsafe set: \{UNSAFE\_SET\}

\vspace{1.5ex}

\textcolor{promptheader}{\textbf{Design a barrier certificate B(x) that satisfies:}}\\
1. B(x) $\leq$ 0 in initial set\\
2. B(x) > 0 in unsafe set\\
3. \{CONDITION\_3\}

\vspace{1.5ex}

Learn from previous failures. You can change structure of TEMPLATE if needed. In this step, the goal is to improve the structure of the templates, not refine the parameters.\\

\vspace{1.5ex}

\textcolor{red}{\textbf{CRITICAL:}} Use ONLY real numbers in the barrier expression. No variables like 'c' or '$\varepsilon$'. Solve specifically for THIS problem with appropriate coefficients.

\vspace{1.5ex}

Analyze carefully but be concise. Give precise answer without long explanations.

\vspace{1.5ex}

\textcolor{promptheader}{\textbf{Format your response as (don't make it bold):}}

\vspace{1.5ex}

BARRIER: [expression with numbers only]
\end{promptbox}

The variable \texttt{FAILED\_INFO} refers to

\begin{center}
   failed: FAILED\_CONDITIONS (number of counter-examples).
\end{center}

\subsubsection{Systems With Control Inputs}
\begin{promptbox}[Barrier and Controller Synthesis in Subsequent Iterations]
\textcolor{promptheader}{\textbf{Previous barrier + controller attempts failed:}}\\
- Barrier: \{BARRIER\}, Controller: \{CONTROLLER\}, \{FAILED\_INFO\}\\
- Barrier: \{BARRIER\}, Controller: \{CONTROLLER\}, \{FAILED\_INFO\}\\
- Barrier: \{BARRIER\}, Controller: \{CONTROLLER\}, \{FAILED\_INFO\}

\vspace{1.5ex}

\textit{Improve the barrier + controller structure to satisfy all conditions.}

\vspace{1.5ex}

\textcolor{promptheader}{\textbf{Main Problem:}}\\
- Dynamics: \{SYSTEM\_DYNAMICS\}\\
- Initial set: \{INITIAL\_SET\}\\
- Unsafe set: \{UNSAFE\_SET\}

\vspace{1.5ex}

\textcolor{promptheader}{\textbf{CONTROLLER SYNTHESIS:}} \\
This system has control inputs
\{CONTROLLER\_PARAMETERS\}. You need to \\ design BOTH:\\
1. Barrier certificate B(x)\\
2. Controller expressions for \{CONTROLLER\_PARAMETERS\}

\vspace{1.5ex}

The controller u(x) will be substituted into dynamics to create closed-loop system.

\vspace{1.5ex}

\textcolor{promptheader}{\textbf{Design a barrier certificate B(x) that satisfies:}}\\
1. B(x) $\leq$ 0 in initial set\\
2. B(x) > 0 in unsafe set\\
3. \{CONDITION\_3\}

\vspace{1.5ex}

Learn from previous failures. You can change structure of TEMPLATE if needed. In this step, the goal is to improve the structure of the templates, not refine the parameters.

\vspace{1.5ex}

Design both barrier certificate B(x) AND controller expressions that work\\
together to satisfy all conditions.

\vspace{1.5ex}

\textcolor{red}{\textbf{CRITICAL:}}\\
- Use ONLY real numbers in both barrier and controller expressions.\\
  No variables like 'c' or '$\varepsilon$'.\\
- Solve specifically for THIS problem with appropriate coefficients.\\
- Controller must be implementable with realistic actuators\\
- Ensure controller bounds are reasonable (avoid extremely large values)

\vspace{1.5ex}

Analyze carefully but be concise. Give precise answer without long explanations.

\vspace{1.5ex}

\textcolor{promptheader}{\textbf{Format your response as (don't make it bold):}}

\vspace{1.5ex}

BARRIER: [barrier expression with numbers only]\\
CONTROLLER: [controller expressions for each parameter, comma-separated]
\end{promptbox}

\subsection{Prompt 6: Coefficient Refinement in the First Two Iterations}

For the first two iterations of the refinement, the prompt focuses on the adjustment of the coefficient without structural changes.

\subsubsection{Systems Without Controller}

\begin{promptbox}[Coefficient Refinement - No Controller]
Original barrier: \{BARRIER\} \{FAILED\_INFO\}
\vspace{1.5ex}

\{REFINEMENT\_HISTORY\}

\vspace{1.5ex}

\textcolor{promptheader}{\textbf{Problem:}}\\
- Dynamics: \{SYSTEM\_DYNAMICS\}\\
- Initial set: \{INITIAL\_SET\}\\
- Unsafe set: \{UNSAFE\_SET\}

\vspace{1.5ex}

Try a different coefficient distribution. You can redistribute the coefficients
between ALL terms (sometimes it is necessary for all terms to have different
coefficients), but DO NOT change structure

\vspace{1.5ex}

\textcolor{promptheader}{\textbf{Requirements:}}\\
1. B(x) $\leq$ 0 in initial set\\
2. B(x) > 0 in unsafe set\\
3. \{CONDITION\_3\}

\vspace{1.5ex}

Analyze carefully but be concise. Give precise answer without long explanations.

\vspace{1.5ex}

\textcolor{promptheader}{\textbf{Format your response as (don't make it bold):}}

\vspace{1.5ex}

REFINED\_BARRIER: [expression with numbers only]
\end{promptbox}

\subsubsection{Systems With Control Inputs}

\begin{promptbox}[Coefficient Refinement - Barrier and Controller]
Original barrier: \{BARRIER\}, Original controller: \{CONTROLLER\}, 
 \{FAILED\_INFO\}
\vspace{1.5ex}

\{REFINEMENT\_HISTORY\}

\vspace{1.5ex}

\textcolor{promptheader}{\textbf{Problem:}}\\
- Dynamics: \{SYSTEM\_DYNAMICS\}\\
- Initial set: \{INITIAL\_SET\}\\
- Unsafe set: \{UNSAFE\_SET\}

\vspace{1.5ex}

Try a different coefficient distribution for both barrier and controller. \\ You can
redistribute the coefficients between ALL terms, but DO NOT change structure

\vspace{1.5ex}

\textcolor{promptheader}{\textbf{CONTROLLER SYNTHESIS CONSTRAINTS:}}\\
1. Controller parameters: \{CONTROLLER\_PARAMETERS\}\\
2. Use smooth, bounded functions (avoid extremely large values)\\
3. Controller must work harmoniously with the barrier\\
4. Ensure closed-loop stability

\vspace{1.5ex}

\textcolor{promptheader}{\textbf{Requirements:}}\\
1. B(x) $\leq$ 0 in initial set\\
2. B(x) > 0 in unsafe set\\
3. \{CONDITION\_3\}

\vspace{1.5ex}

Analyze carefully but be concise. Give precise answer without long explanations.

\vspace{1.5ex}

\textcolor{promptheader}{\textbf{Format your response as (don't make it bold):}}

\vspace{1.5ex}

REFINED\_BARRIER: [barrier expression with numbers only]\\
REFINED\_CONTROLLER: [controller expressions for each parameter, comma-separated]
\end{promptbox}

\subsection{Prompt 7: Structure Refinement in Subsequent Iterations}

For refinement iterations 3 and 4, the prompt allows structural modifications when coefficient adjustments fail.

\subsubsection{Systems Without Controller}

\begin{promptbox}[Structure Refinement - No Controller]
Original barrier: \{BARRIER\}  \{FAILED\_INFO\}

\vspace{1.5ex}

\{REFINEMENT\_HISTORY\}

\vspace{1.5ex}

\textcolor{promptheader}{\textbf{Problem:}}\\
- Dynamics: \{SYSTEM\_DYNAMICS\}\\
- Initial set: \{INITIAL\_SET\}\\
- Unsafe set: \{UNSAFE\_SET\}

\vspace{1.5ex}

Previous coefficient adjustments failed. Consider changing the barrier structure
if needed while keeping the same polynomial degree. You can modify the terms or
their combinations.

\vspace{1.5ex}

\textcolor{promptheader}{\textbf{Requirements:}}\\
1. B(x) $\leq$ 0 in initial set\\
2. B(x) > 0 in unsafe set\\
3. \{CONDITION\_3\}

\vspace{1.5ex}

Analyze carefully but be concise. Give precise answer without long explanations.

\vspace{1.5ex}

\textcolor{promptheader}{\textbf{Format your response as (don't make it bold):}}

\vspace{1.5ex}

REFINED\_BARRIER: [expression with numbers only]
\end{promptbox}

\subsubsection{Systems With Control Input}

\begin{promptbox}[Structure Refinement - Barrier and Controller]
Original barrier: \{BARRIER\}, Original controller: \{CONTROLLER\},
 \{FAILED\_INFO\}

\vspace{1.5ex}

\{REFINEMENT\_HISTORY\}

\vspace{1.5ex}

\textcolor{promptheader}{\textbf{Problem:}}\\
- Dynamics: \{SYSTEM\_DYNAMICS\}\\
- Initial set: \{INITIAL\_SET\}\\
- Unsafe set: \{UNSAFE\_SET\}

\vspace{1.5ex}

Previous coefficient adjustments failed. Consider changing the barrier and/or
controller structure if needed while keeping the same polynomial degree. You can
modify the terms or their combinations.

\vspace{1.5ex}

\textcolor{promptheader}{\textbf{CONTROLLER SYNTHESIS CONSTRAINTS:}}\\
1. Controller parameters: \{CONTROLLER\_PARAMETERS\}\\
2. Use smooth, bounded functions (avoid extremely large values)\\
3. Controller must work harmoniously with the barrier\\
4. Ensure closed-loop stability

\vspace{1ex}

\textcolor{promptheader}{\textbf{Requirements:}}\\
1. B(x) $\leq$ 0 in initial set\\
2. B(x) > 0 in unsafe set\\
3. \{CONDITION\_3\}

\vspace{1ex}

Analyze carefully but be concise. Give precise answer without long explanations.

\vspace{1ex}

\textcolor{promptheader}{\textbf{Format your response as (don't make it bold):}}

\vspace{1ex}

REFINED\_BARRIER: [barrier expression with numbers only]\\
REFINED\_CONTROLLER: [controller expressions for each parameter, comma-separated]
\end{promptbox}

\textit{Refinement History Format:}
\begin{lstlisting}[style=promptstyle, basicstyle=\ttfamily\scriptsize]
Refinement 1: {BARRIER} {FAILED_INFO}
Refinement 2: {BARRIER} {FAILED_INFO}
...
\end{lstlisting}

\subsection{Prompt 8: SMT Solver}

\subsubsection{Solver Selection}

This prompt selects the most appropriate SMT solver for verification.

\begin{promptbox}[SMT Solver Selection]
Select the best SMT solver based on barrier expression and the dynamical system.

\vspace{1.5ex}

\textcolor{promptheader}{\textbf{PROBLEM:}}\\
Dynamics: \{SYSTEM\_DYNAMICS\}\\
Barrier: \{BARRIER\_EXPRESSION\}

\vspace{1.5ex}

\textcolor{promptheader}{\textbf{AVAILABLE SOLVERS:}}\\
- cvc5\\
- z3\\
- yices

\vspace{1.5ex}

Without long explanations, give a short and precise answer.

\vspace{1.5ex}

\textcolor{promptheader}{\textbf{Format your response as:}}

\vspace{1.5ex}

SOLVER: [solver name]
\end{promptbox}

\subsubsection{Timeout Analysis}

This prompt determines whether to retry with extended timeout when a solver times out.

\begin{promptbox}[SMT Solver Timeout Analysis]
Solver \{SOLVER\_NAME\} timed out after \{TIMEOUT\_MS\}ms.

\vspace{1.5ex}

\textcolor{promptheader}{\textbf{PROBLEM:}}\\
Dynamics: \{SYSTEM\_DYNAMICS\}\\
Barrier: \{BARRIER\_EXPRESSION\}

\vspace{1.5ex}

Given the above information, should we attempt again with more time, or is this
barrier too complex to verify?

\vspace{1.5ex}

Without long explanations, give a short and precise answer.

\vspace{1.5ex}

\textcolor{promptheader}{\textbf{Format your response as:}}

\vspace{1.5ex}

RETRY: yes or no\\
TIMEOUT\_MULTIPLIER: [number] (only if RETRY is yes, e.g., 1.5 or 2.0)
\end{promptbox}

\subsubsection{Error Analysis}

This prompt selects an alternative solver when verification fails.

\begin{promptbox}[Alternative Solver Selection]
Solver \{SOLVER\_NAME\} failed during verification.

\vspace{1.5ex}

ERROR: \{ERROR\_TYPE\}\\
MESSAGE: \{ERROR\_MESSAGE\}

\vspace{1.5ex}

\textcolor{promptheader}{\textbf{PROBLEM:}}\\
Dynamics: \{SYSTEM\_DYNAMICS\}\\
Barrier: \{BARRIER\_EXPRESSION\}

\vspace{1.5ex}

\textcolor{promptheader}{\textbf{REMAINING SOLVERS:}}\\
\{REMAINING\_SOLVERS\_LIST\}

\vspace{1.5ex}

Select a different solver to try.

\vspace{1.5ex}

Without long explanations, give a short and precise answer.

\vspace{1.5ex}

\textcolor{promptheader}{\textbf{Format your response as:}}

\vspace{1.5ex}

NEXT\_SOLVER: [solver name]
\end{promptbox}

\end{document}